\documentclass[sigconf]{acmart}

\setcopyright{none}              
\renewcommand\footnotetextcopyrightpermission[1]{} 
\AtBeginDocument{%
  }

\begin{document}

\title{SHARE: Scene-Human Aligned Reconstruction}

\author{Joshua Li}
\affiliation{%
  \institution{National Research Council Canada}
  \country{Canada}
}
\affiliation{%
  \institution{University of Waterloo}
  \country{Canada}
}
\email{j234li@uwaterloo.ca}

\author{Brendan Chharawala}
\affiliation{%
  \institution{National Research Council Canada}
  \country{Canada}
}
\affiliation{%
  \institution{University of Waterloo}
  \country{Canada}}
\email{bchharaw@uwaterloo.ca}

\author{Chang Shu}
\affiliation{%
  \institution{National Research Council Canada}
  \country{Canada}}
\email{chang.shu@nrc-cnrc.gc.ca}

\author{Xue Bin Peng}
\affiliation{%
  \institution{Simon Fraser University}
  \country{Canada}}
\email{xbpeng@sfu.ca}

\author{Pengcheng Xi}
\affiliation{%
  \institution{National Research Council Canada}
  \country{Canada}}
\email{pengcheng.xi@nrc-cnrc.gc.ca}

\renewcommand{\shortauthors}{Li et al.}

\begin{abstract}
  Animating realistic character interactions with the surrounding environment is important for autonomous agents in gaming, AR/VR, and robotics. However, current methods for human motion reconstruction struggle with accurately placing humans in 3D space. We introduce Scene-Human Aligned REconstruction (SHARE), a technique that leverages the scene geometry's inherent spatial cues to accurately ground human motion reconstruction. Each reconstruction relies solely on a monocular RGB video from a stationary camera. SHARE first estimates a human mesh and segmentation mask for every frame, alongside a scene point map at keyframes. It iteratively refines the human's positions at these keyframes by comparing the human mesh against the human point map extracted from the scene using the mask. Crucially, we also ensure that non-keyframe human meshes remain consistent by preserving their relative root joint positions to keyframe root joints during optimization. Our approach enables more accurate 3D human placement while reconstructing the surrounding scene, facilitating use cases on both curated datasets and in-the-wild web videos. Extensive experiments demonstrate that SHARE outperforms existing methods.
\end{abstract}

\ccsdesc[500]{Computing methodologies~Reconstruction}
\ccsdesc[500]{Computing methodologies~Motion capture}

\begin{teaserfigure}
  \centering
  \includegraphics[width=0.95\textwidth]{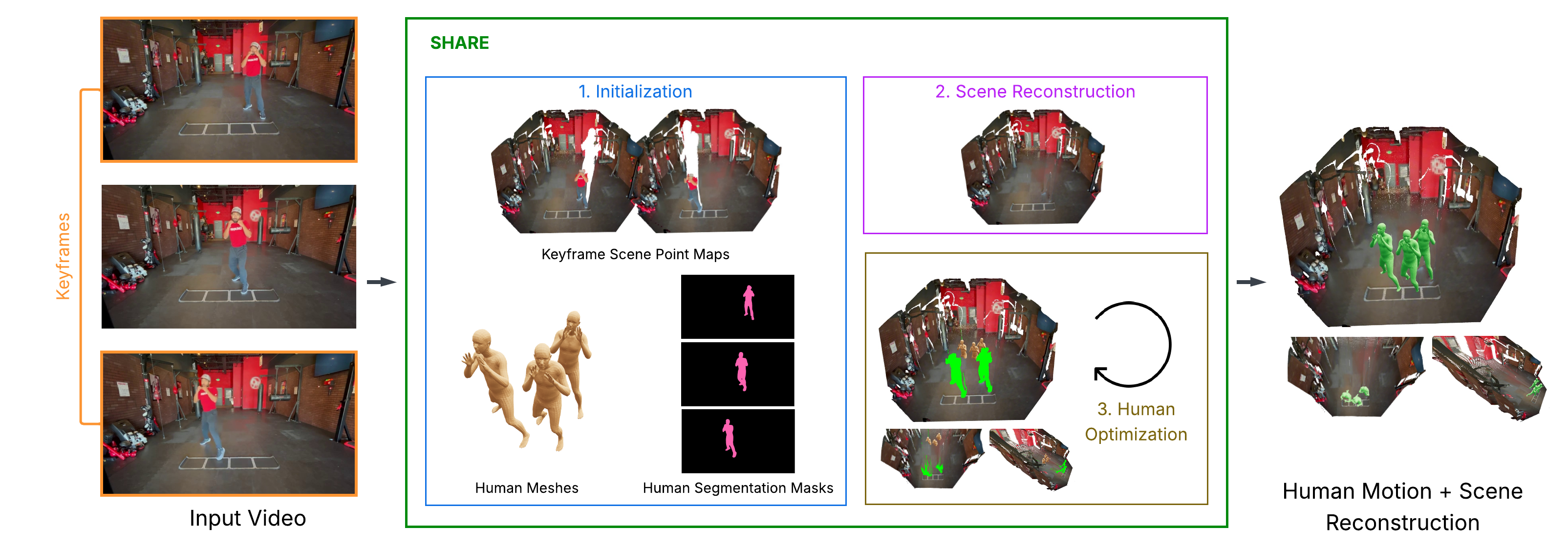}
  \caption{\textbf{Overview of SHARE}. Our framework takes a video as input and operates in three stages: (1) initialize point maps, human meshes and masks using pre-trained models; (2) reconstruct the background scene; (3) optimize the human meshes by grounding them to human scene points. The output of SHARE is a reconstruction of both the human motion and the scene in metric scale. \textit{Video credit: \cite{TheRealDealMMATips2025}. Identifying faces and logos are blurred for privacy.}}
  \Description{Overview of SHARE.}
  \label{fig:teaser}
\end{teaserfigure}

\maketitle
\fancyhead{}      
\fancyfoot{}      
\thispagestyle{plain}  
\section{Introduction}
Creating realistic motion for virtual characters enhances user immersion in gaming and AR/VR, and enables more natural human-robot collaboration. Yet human motion rarely occurs in isolation -- we constantly interact with our environment, whether it is navigating around a room, sitting on a sofa, or lying on a bed.
Therefore, generating plausible motion demands grounding it to the surrounding scene.

To address this, recent methods have focused on explicitly incorporating Human-Scene Interaction (HSI) into motion generation. Data-driven methods learn motion patterns directly from paired human-scene datasets; however, such data is difficult to obtain. Real-world captures are costly and limited in scale, while synthetic datasets often lack diverse interactions. Alternatively, reinforcement learning approaches can leverage unpaired human and scene data, but depend on carefully crafted objective functions.

Human motion reconstruction from monocular videos is a well-established task and presents a promising pathway toward joint human-scene understanding. However, approaches that recover both human motion and the surrounding environment remain relatively underexplored. Existing methods also face notable limitations: MHMocap \cite{SceneAware_EG2023} depends on multiple loss functions, while JOSH \cite{liu2025jointoptimization4dhumanscene} requires a separate model to predict contacts. 

We introduce Scene-Human Aligned REconstruction (SHARE), which achieves accurate results without these limitations. SHARE uses two straightforward loss functions and directly exploits the strong depth cues from point maps to ground the predicted human meshes. In summary, our contributions are twofold: (1) we propose a novel human motion reconstruction approach guided by estimated scene point maps, which demonstrates improved 3D human positioning; and (2), we introduce a unified framework that reconstructs both human motion and the surrounding scene, enabling data-driven motion generation for HSI.

\section{Method}
\label{sec:method}

\newcommand{\colorBox}[1]{%
  \raisebox{0.5ex}{%
    \fcolorbox{black}{#1}{%
      \rule{0pt}{0.5ex}\hspace{0.5ex}%
    }%
  }%
}
\definecolor{gt}{RGB}{255, 255, 255}
\definecolor{mhmocap}{RGB}{153, 199, 255}
\definecolor{tram}{RGB}{255, 169, 77}
\definecolor{bev}{RGB}{255, 115, 169}
\definecolor{share}{RGB}{59, 255, 65}

\begin{figure*}[h]
  \centering
  \includegraphics[width=0.95\linewidth]{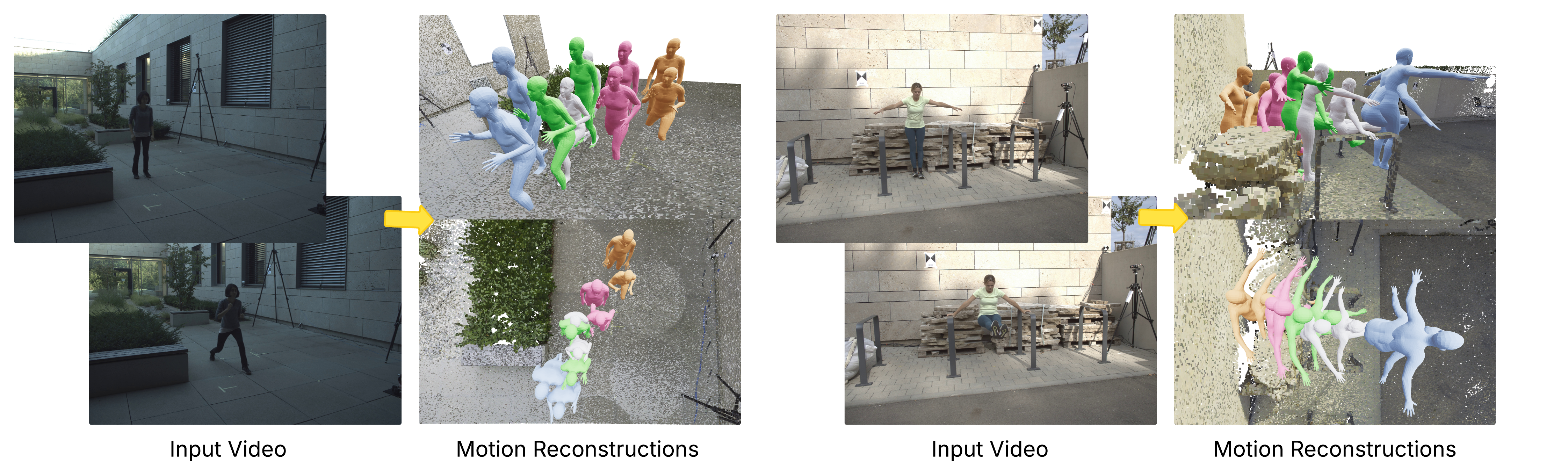}
  \caption{\textbf{Comparison of human motion reconstruction methods}. Ground truth RICH motion is shown in \colorBox{gt}, MHMocap in \colorBox{mhmocap}, TRAM in \colorBox{tram}, BEV in \colorBox{bev}, and SHARE in \colorBox{share}. We place all reconstructions in ground truth scene scans. Our method, SHARE, achieves the closest alignment to the ground truth motion.}
  \Description{Comparison of human motion reconstruction methods.}
  \label{fig:rich}
\end{figure*}

Our method assumes a video from a stationary camera with a single person in view. Given an RGB video, represented as a set of frames $\{f_t \in \mathbb{R}^{h \times w \times 3}\}_{t=1}^{T}$, we aim to reconstruct a 3D mesh of the person at every frame. We do so by estimating the parameters for the Skinned Multi-Person Linear Model (SMPL) \cite{SMPL:2015}, a differentiable model that generates a realistic human mesh from the local pose $\theta \in \mathbb{R}^{24 \times 3}$, body shape $\beta \in \mathbb{R}^{10}$, and translation $\pi \in \mathbb{R}^3$. In addition to the human mesh, we also reconstruct a metric-scale background scene point map $S \in \mathbb{R}^{p \times 6}$, where $p$ is the number of valid pixels and each point is represented by XYZRGB (position and color). An overview of our framework is illustrated in Fig. \ref{fig:teaser}.

\subsection{Initialization}
We separately initialize human meshes and segmentation masks using TRAM \cite{wang2024tram}, and keyframe scene point maps using MoGe-2 \cite{wang2025moge2accuratemonoculargeometry}. Both models deliver strong performance in their respective tasks.

TRAM is a method for recovering human motion from any video. Although we do not utilize TRAM's SLAM in our stationary setting, we leverage its robust human masking component. This pipeline integrates human detection, segmentation, and tracking to produce human masks $\big\{H_t \in \{0, 1\}^{h \times w}\big\}_{t=1}^{T}$. For simplicity, a mask can be equivalently represented as a set $H_t \subset \Omega$, where $\Omega$ is the set of all pixels. Additionally, we use TRAM's VIMO transformer model to estimate initial SMPL parameters, which include local poses $\{\theta_t\}_{t=1}^{T}$, a single body shape $\beta$ and translations $\{\pi_t\}_{t=1}^{T}$.

MoGe-2 is a geometry estimation model that recovers a 3D point map from an image. This model improves upon its predecessor, MoGe, by enhancing its ViT-based architecture with metric-scale prediction and improved training data. We generate a 3D point map by projecting the model’s estimated 2D depth map using its predicted camera intrinsics, with each 3D point inheriting the color from its corresponding pixel in the original image. To ensure computational efficiency, we also process a sparse set of keyframes rather than the entire video. For shorter clips, we found that using just the first and last frames (frame $1$ and frame $T$) is sufficient, which provides us with depth maps $D_1, D_{T} \in \mathbb{R}^{h \times w}$ and camera intrinsics $I_1, I_{T} \in \mathbb{R}^{3 \times 3}$. While incorporating additional keyframes can enhance performance, it comes at the cost of reduced efficiency. Note that we always implicitly apply MoGe-2's built-in mask to filter out noisy points from the depth map.

\subsection{Scene Reconstruction}
MoGe-2 can produce slightly different depth maps for the same scene across different frames. To align them, we scale the depth map from the second keyframe to match the first. We calculate a scale factor $\alpha$ by dividing the mean depth of $D_{1}$ by the mean depth of $D_T$  for all pixels outside the human masks, i.e. $i \in \Omega \setminus \{H_1 \cup H_T\}$. The aligned 3D point maps $S_1$, $S_T$ are then obtained using the inverse projection function $\Pi^{-1}$ as follows:
\begin{equation}
    S_1=\Pi^{-1}(D_1, I_1, f_1)
\end{equation}
\begin{equation}
    S_T=\alpha \cdot \Pi^{-1}(D_T, I_T, f_T)
\end{equation}
We then reconstruct a single scene point map by combining the point maps from both keyframes, excluding the human pixels. We operate under the strong assumption that the human masks in the keyframes do not overlap, which allows for a complete reconstruction of the background. To create the final scene point map, $S$, we take the average of the common background points from both keyframes. We then add any points that were under the human mask in one keyframe from the complementary keyframe. This process is summarized by the following equation:
\begin{multline}
S=\bigg\{ \Big\{\frac{1}{2} (S_1^i+S_{T}^i): i \in \Omega \setminus \{H_1 \cup H_{T} \} \Big\} \\ 
\cup \{ S_1^i :i \in H_{T}\}
\cup \{S_{T}^i: i \in H_1\} \bigg\}
\end{multline}
Here, $S$ uses the assumption of a stationary camera to directly reconstruct the scene by merging keyframe points.

\subsection{Human Optimization}
While MoGe-2 provides a good metric scale for the human's position within the scene, it does not accurately capture fine body shape details. Therefore, we freeze the local poses $\{\theta_t\}_{t=1}^{T}$ and body shape $\beta$ predicted by TRAM and only optimize for the human translations $\{\pi_t\}_{t=1}^{T}$. The SMPL model takes these parameters to produce body joints $\{J_t \in \mathbb{R}^{24 \times 3}\}_{t=1}^{T}$ and human mesh vertices $\{V_t \in \mathbb{R}^{6890 \times 3}\}_{t=1}^{T}$. Since SMPL is differentiable, we can use gradient descent to minimize a loss $\mathcal{L}$:
\begin{equation}
    \mathcal{L}=\mathcal{L}_{body}+\mathcal{L}_{root}
\end{equation}
In brief, $\mathcal{L}_{body}$ anchors the human mesh to the human point map at keyframes and $\mathcal{L}_{root}$ adjusts the non-keyframe meshes accordingly.

\textbf{Keyframe Body Loss}.
We calculate the symmetric Chamfer Distance ($CD$) loss between the human mesh vertices and the human point map at both keyframes:
\begin{equation}
    \mathcal{L}_{body}=CD(V_1, \{S_1^i: i \in H_1\})+CD(V_{T}, \{S_T^i: i \in H_T\})
\end{equation}
This loss encourages the keyframe translations $\pi_1, \pi_T$ to robustly place the human meshes within the densest part of the noisy point maps.

\textbf{Relative Root Joint Loss}.
The original TRAM method reduces motion jitter by applying a Gaussian filter to the root joint trajectory during a post-processing step. We aim to achieve this smoothing directly by optimizing the translation parameters. Additionally, we would like to ensure that any adjustment to a keyframe's body position by the $\mathcal{L}_{body}$ loss causes the entire motion trajectory to shift in a consistent manner. 

First, we use a Gaussian filter $\mathcal{G}$ to get the smoothed root joint positions before optimization:

\begin{equation}
    \tilde{J}^{root}_{init} = \mathcal{G}(J^{root}_{init})
\end{equation}

Next, we calculate the root joint positions relative to each keyframe for both the smoothed initial motion and the current motion, while ensuring the keyframes are not directly tied to each other. For a keyframe $k$ and its complementary keyframe $k'$, where $(k, k') \in \{(1, T), (T, 1)\}$, we define:

\begin{equation}
    d_{init, k}=\Big[\tilde{J}^{root}_{init, t} - \tilde{J}^{root}_{init, k} \Big]_{t \in \{ 1...T \} \setminus k'}
\end{equation}
\begin{equation}
    d_k=\Big[J^{root}_t - J^{root}_k\Big]_{t \in \{ 1...T \} \setminus k'}
\end{equation}

Finally, we minimize the following mean squared error loss:
\begin{equation}
    \mathcal{L}_{root}=\frac{1}{T-1}(||d_{init,1}-d_1||_2^2 + ||d_{init,T}-d_{T}||_2^2)
\end{equation}
This loss encourages the current motion to follow the initial smoothed trajectory.
\section{Results}

\begin{table}
  \caption{\textbf{Human motion reconstruction results}. We evaluate on a 3.3k-frame subset of the RICH-test dataset. The "Scene" column indicates whether a method also reconstructs the 3D scene. Results reported in mean $\pm$ std.}
  \label{tab:freq}
  \begin{tabular}{p{3.3cm}ccc}
    \toprule
    Method & Scene & MRPE (m) $\downarrow$ & V2V (m) $\downarrow$\\
    \midrule
    MHMocap \cite{SceneAware_EG2023} & $\checkmark$ & 1.19 $\pm$ 0.67 & 1.20 $\pm$ 0.67 \\
    TRAM \cite{wang2024tram} & $\times$ & 0.98 $\pm$ 0.37 & 0.98 $\pm$ 0.36 \\
    BEV \cite{BEV}& $\times$ & 0.61 $\pm$ 0.35 & 0.62 $\pm$ 0.33 \\
    SHARE (ours) & $\checkmark$ & \textbf{0.44 $\pm$ 0.37} & \textbf{0.44 $\pm$ 0.36} \\
  \bottomrule
\end{tabular}
\end{table}

\subsection{Human Motion Reconstruction}

\textbf{Dataset}. We evaluate our method on a subset of the RICH dataset's test split \cite{Huang:CVPR:2022}. This subset, totaling 3.3k frames, consists of 33 100-frame sequences derived from 8 videos across 3 scenes. All sequences feature a single person and a static camera, and were specifically chosen to ensure every compared model could produce valid results.

\textbf{Evaluation metrics}. Following previous works \cite{Huang:CVPR:2022, SceneAware_EG2023}, we use two metrics for evaluation: Mean Root Position Error (MRPE) and Vertex-to-Vertex Error (V2V). MRPE gauges the accuracy of the pelvis placement, while V2V measures the overall fidelity of the reconstructed mesh.

\textbf{Implementation details}. Our method is implemented using PyTorch. We run 600 optimization iterations using an Adam optimizer with a learning rate of 0.01. All computations are performed on a single NVIDIA RTX A6000 GPU with 48 GB of memory.

\textbf{Quantitative Results}. As shown in Table \ref{tab:freq}, SHARE outperforms all prior works on both metrics by a sizable margin. This performance can be attributed to our method's enhanced accuracy in predicting human body translations in 3D space. MHMocap is the only other SMPL-based method that also reconstructs the scene from an RGB video with a stationary camera, but its results are limited by inaccuracies from its depth estimation model, DPT \cite{DPT}. In contrast, TRAM and BEV reconstruct human motion without accounting for the scene at all. Visualizations of these methods on selected RICH dataset videos are available in Fig. \ref{fig:rich}.

\subsection{Human Motion + Scene Reconstruction}
In addition to our quantitative results on the RICH dataset, we emphasize SHARE's generalization across diverse use cases. For example, Fig. \ref{fig:teaser} shows its strong qualitative performance on an in-the-wild web video. 

\begin{figure}[t]
  \centering
  \includegraphics[width=\linewidth]{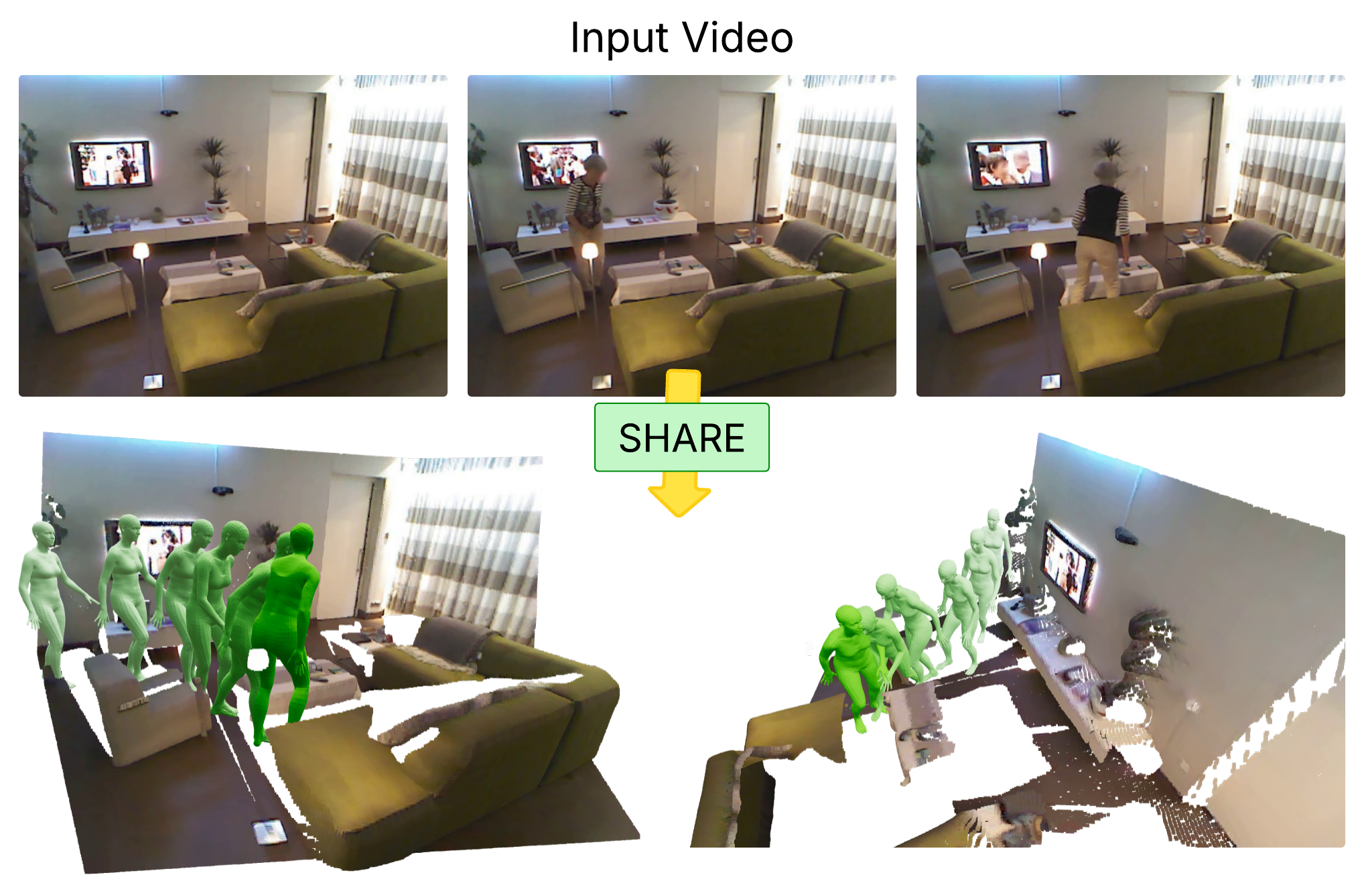}
  \caption{\textbf{Human motion + scene reconstruction using SHARE}. We showcase reconstruction on a video of an older adult walking around a table in the Toyota Smarthome dataset.}
  \Description{Human motion + scene reconstruction using SHARE.}
  \label{fig:toyota}
\end{figure}

Moreover, SHARE effectively reconstructs human motion and scene in curated datasets that lack ground truth human motion or scene annotations. To showcase this capability, we applied SHARE to the Toyota Smarthome dataset \cite{toyota_smarthome}, which captures the daily activities of older adults. While the dataset provides clean RGB videos, it suffers from noisy depth and skeletal data. Fig. \ref{fig:toyota} shows an example of our reconstruction, which produces more reliable data modalities.

We further validate our method through a user study involving 20 participants. From the Toyota Smarthome dataset, we randomly selected a subset of videos featuring walking motions, processed them using MHMocap and SHARE, and then randomly chose 3 pairs of successful reconstructions. Participants rated the realism of each reconstruction on a scale of 1 to 5 and indicated their preferred reconstruction in each pair. The results, summarized in Table \ref{tab:user}, reveal a clear preference for SHARE. We argue that SHARE's ability to reconstruct human motion and scene makes it especially useful for researchers aiming to extract HSI motions from in-the-wild web videos or curated datasets without relying on existing annotations.

\begin{table}
  \caption{\textbf{User study results}. 20 participants evaluated the realism of human and scene reconstructions across 3 videos from the Toyota Smarthome dataset.}
  \label{tab:user}
  \begin{tabular}{lcc}
    \toprule
    Method & Score & Selection Rate\\
    \midrule
    MHMocap \cite{SceneAware_EG2023} & 2.07 & 15\% \\
    SHARE (ours) & \textbf{3.22} & \textbf{85\%} \\
  \bottomrule
\end{tabular}
\end{table}
\section{Conclusion}
We introduce SHARE, a framework that reconstructs both human motion and the surrounding environment from monocular videos. Quantitative results demonstrate that our approach yields significant improvements in 3D positional accuracy for human motion reconstruction. We anticipate that SHARE can be applied to a wide range of video data and envision extensions towards physically plausible motion reconstruction for human-scene interaction.

\textbf{Limitations}. Our method assumes input from a stationary camera observing a single human, which limits its applicability to scenarios involving camera motion or multiple individuals. Additionally, the human reconstruction accuracy is dependent on the quality of initialization from TRAM and MoGe-2. For example, mismatches between the human mesh size and the human point map height can lead to artifacts such as floor penetration. Finally, during optimization, discrepancies between the initial smoothed keyframe meshes and the current ones may also introduce jitter at keyframes.

\bibliographystyle{ACM-Reference-Format}
\bibliography{citations}

\end{document}